%% file: main_1.1.tex
\documentclass[nohyperref]{article}

\usepackage{microtype}
\usepackage{graphicx}
\usepackage{subfigure}
\usepackage{booktabs} % for professional tables

\usepackage{hyperref}

\input{math_commands.tex}

\usepackage[accepted]{icml2022}

\usepackage{amsmath}
\usepackage{amssymb}
\usepackage{mathtools}
\usepackage{amsthm}
\usepackage{mathrsfs}
\usepackage{amsfonts}
\usepackage[capitalize,noabbrev]{cleveref}

\theoremstyle{plain}

\theoremstyle{definition}

\theoremstyle{remark}

\usepackage[textsize=tiny]{todonotes}

\icmltitlerunning{Penalizing Gradient Norm for Efficiently Improving Generalization in Deep Learning}

\begin{document}

\twocolumn[
\icmltitle{Penalizing Gradient Norm for Efficiently Improving \\ Generalization in Deep Learning}



\icmlsetsymbol{equal}{*}

\begin{icmlauthorlist}
\icmlauthor{Yang Zhao}{thu}
\icmlauthor{Hao Zhang}{thu}
\icmlauthor{Xiuyuan Hu}{thu}
\end{icmlauthorlist}

\icmlaffiliation{thu}{Department of Electronic Engineering, Tsinghua University}
\icmlcorrespondingauthor{Hao Zhang}{haozhang@tsinghua.edu.cn}
\icmlcorrespondingauthor{Yang Zhao}{zhao-yan18@mails.tsinghua.edu.cn}

\icmlkeywords{Machine Learning, ICML}

\vskip 0.3in
]



\printAffiliationsAndNotice{}  % leave blank if no need to mention equal contribution

\begin{abstract}

% How to train deep neural networks (DNNs) to generalize well is a central concern in deep learning, especially for severely overparameterized networks nowadays. In this paper, we propose an effective method to improve the model generalization by additionally penalizing the gradient norm of loss function during optimization. We demonstrate that confining the gradient norm of loss function could help lead the optimizers towards finding flat minima. We leverage the first-order approximation to efficiently implement the corresponding gradient in the gradient descent framework. In our experiments, we confirm that when using our methods, generalization performance of various models could be improved on different datasets. Also, we show that the recent sharpness-aware minimization method is a special, but not the best, case of our method, where the best case of our method could give new state-of-art performance on these tasks.

How to train deep neural networks (DNNs) to generalize well is a central concern in deep learning, especially for severely overparameterized networks nowadays. In this paper, we propose an effective method to improve the model generalization by additionally penalizing the gradient norm of loss function during optimization. We demonstrate that confining the gradient norm of loss function could help lead the optimizers towards finding flat minima. We leverage the first-order approximation to efficiently implement the corresponding gradient to fit well in the gradient descent framework. In our experiments, we confirm that when using our methods, generalization performance of various models could be improved on different datasets. Also, we show that the recent sharpness-aware minimization method \cite{DBLP:conf/iclr/ForetKMN21} is a special, but not the best, case of our method, where the best case of our method could give new state-of-art performance on these tasks. Code is available at \href{https://github.com/zhaoyang-0204/gnp}{https://github.com/zhaoyang-0204/gnp}.

\end{abstract}

\section{Introduction}

Today's powerful computation hardwares make it possible for training large-scale deep neural networks (DNNs) \cite{DBLP:journals/corr/GoyalDGNWKTJH17,DBLP:conf/cvpr/HanKK17,DBLP:conf/iclr/DosovitskiyB0WZ21}. These DNNs typically have millions or even billions of parameters, completely far exceeding the amount of training samples. Due to such heavy parametrization, they are capable to provide larger hypothesis space with normally better solutions. But in the meantime, such a huge hypothesis space is also full of more minima with diverse generalization ability \cite{DBLP:conf/nips/NeyshaburBMS17}. This makes it more challanging to train them to converge to optimal minima at which models would generalize better. Therefore, how to guide optimizers to find such optimal minima becomes a more salient concern than ever.

Generally, even if the given training datasets have been fully utilized, minimizing only the training loss gauging the gap between the true labels and predicted labels still could not ensure convergence to satisfactory minima. Regarding this, implementing regularization would play a critical role in modern training paradigm \cite{DBLP:conf/icml/IoffeS15,DBLP:journals/jmlr/SrivastavaHKSS14}. Regularization techniques may contribute in various ways beyond datasets. In particular, regularizing models to have certain "good" properties could be one of the most commonly used techniques, typically implemented through penalty function methods \cite{smith1997penalty}.

\begin{figure}[t]
    % \vskip 0.2in
    \begin{center}
    \centerline{\includegraphics[width=0.85\columnwidth]{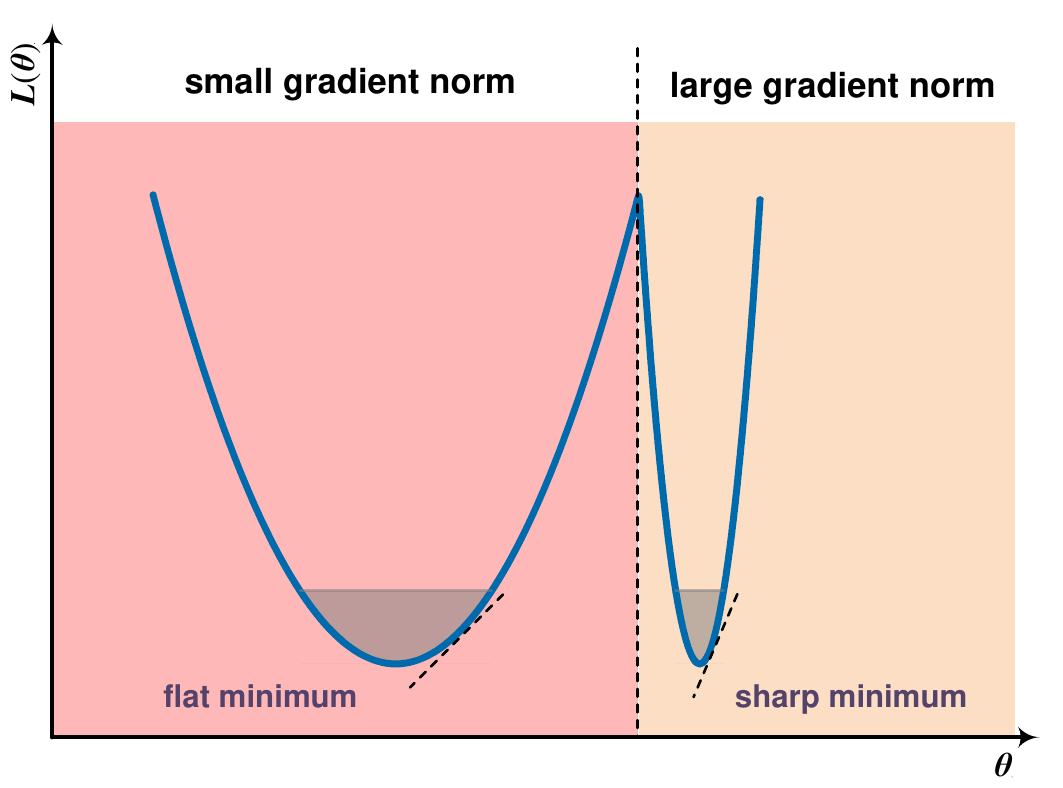}}
    \vskip -0.1in
    \caption{Toy example illustrating connections between the gradient norm of a function and the flatness of the function landscape.}
    \label{fig : introduction}
    \end{center}
    \vskip -0.3in
\end{figure}

In this paper, in addition to optimizing the common loss function, we would further impose an extra penalty on a specific property, the gradient norm of the loss function. The motivation of penalizing the gradient norm of loss function is to encourage the optimizer to find a minimum that lies in a relatively flat neighborhood region, since such flat minima have been demonstrated to be able to lead to better model generalization than sharp ones \cite{DBLP:journals/neco/HochreiterS97a}. \Figref{fig : introduction} gives a toy example that illustrates the association between gradient norm and flatness of minima intuitively, and we would further demonstrate this from the perspective of Lipschitz continuity in \Secref{sec : lip con}.

Unfortunately, for practical implementation, optimizing the gradient norm in a straightforward way would involve the full calculation of Hessian matrix, which is not feasible for current hardwares. Here, by leveraging the approximation techniques, we present a simple and efficient scheme for computing the gradient of this gradient norm. The scheme would avoid the computation of the second-order derivative, and instead use basic algebraic operations between only the first-order derivatives for approximation, thus it could be implemented in practice easily. In particular, we find that the sharpness-aware minimization (SAM) scheme \cite{DBLP:conf/iclr/ForetKMN21} is actually one special case of our scheme, where the hyperparameters are set to specific values.

% Unfortunately, for practical implementation, optimizing the gradient norm in a straightforward way would involve the full calculation of Hessian matrix, which is not feasible for current hardwares. Here, by leveraging the approximation techniques, we present a simple and efficient scheme for computing the gradient of this gradient norm. The scheme would avoid the computation of the second-order derivative, and instead use basic algebraic operations between only the first-order derivatives for approximation, thus it could be implemented in practice easily. 

In our experiments, we apply extensive model architectures on Cifar-\{10, 100\} datasets and ImageNet datasets, respectively. These models would include both simple and complex convolutional neural network architectures, as well as the recent vision transformer architectures. We observe that the model performance could be generally improved via our optimization scheme, and such improvements could be up to 70\% greater than SAM's improvements over the standard training. Also, in some cases, training could be more stable when using our scheme compared to the SAM scheme. Finally, we provide a guide on hyperparameter selection in expectation to achieve the best improvements in practice.

\section{Related Works}

\paragraph{Reguralization techniques}
Regularization could widely refer to techniques that in some way help improve the model generalization, including penalty function methods \cite{smith1997penalty}, data augmentation \cite{DBLP:journals/corr/abs-1708-04552,DBLP:journals/corr/abs-1805-09501}, dropout regularizations \cite{DBLP:journals/jmlr/SrivastavaHKSS14,DBLP:conf/icml/WanZZLF13}, normalization techniques \cite{DBLP:conf/icml/IoffeS15,DBLP:journals/corr/BaKH16,DBLP:conf/eccv/WuH18} and so on. For penalty function methods, extra terms would be added and optimized along with the loss function, which targets to impose constraint on specific property of models. In particular, the weight norm has been demonstrated to be an important property related to the model capacity \cite{DBLP:conf/nips/NeyshaburBMS17}, and penalizing the weight $\normltwo$-norm \cite{DBLP:conf/nips/KroghH91,DBLP:conf/iclr/LoshchilovH19} has become, in a sense, the essential ingredient in modern training recipes. Others like \citet{DBLP:journals/corr/YoshidaM17} penalize the spectral norm of weights for reducing the models' sensitivity to input perturbation.

\paragraph{Flat minima}
On the other hand, our work is also relevant to the research of flat minima. In \citet{DBLP:journals/neco/HochreiterS97a}, the authors first point out that well generalized models may have flat minima. Since then, the association between flatness of minima and model generalization have been studied from both empirical \cite{DBLP:conf/iclr/KeskarMNST17} and theoretical perspectives \cite{DBLP:conf/icml/DinhPBB17,DBLP:conf/nips/NeyshaburBMS17}. Although SGD optimizer and some of its variants (such as momentum) could somehow serve as implicit regularizations that favors flat minima \cite{DBLP:books/daglib/0040158, DBLP:conf/nips/WuME18, DBLP:conf/iclr/XieSS21}, researchers also desire to bias the optimizers in an explicit way in pursuit of smoother surface and flatter minima to further improve model performance, especially for modern scalable models. But in practical optimization, explicitly finding flat minima is nontrivial. Recently, \citet{DBLP:conf/iclr/ForetKMN21} treat it as a minimax optimization problem, and solve it by introducing an efficient procedure, called SAM. Model generalization could be improved significantly compared to using vanilla SGD optimizations. Further, based on \citet{DBLP:conf/iclr/ForetKMN21}, \citet{DBLP:conf/icml/KwonKPC21} propose the Adaptive SAM, where optimization could keep invariant to a specific weight-rescaling operation discussed in \cite{DBLP:conf/nips/NeyshaburSS15,DBLP:conf/icml/DinhPBB17}; \citet{DBLP:conf/cvpr/ZhengZM21} perform gradient descent twice in one step to solve the corresponding minimax problem, one for the inner maximization optimization and the other for the outer minimization optimization. 

\section{Method}

\subsection{Basic Setting}
Given a training dataset $\gS = \{(\vx_i, \vy_i)\}_{i=0}^{n}$ drawn i.i.d from the distribution $\mathscr{D}$, a neural network $f(\cdot; \vtheta)$ is trained to learn this distribution. The neural network is parametrized with parameters $\vtheta$ in weight space $\mathbf{\Theta}$, which would be optimized via minimizing an empirical loss function $L_{\gS}(\vtheta) = \frac{1}{N} \sum_{i=1}^{N} l(\hat{\vy}_i, \vy_i, \vtheta)$ where $\hat{\vy}_i = f(\vx_i;\vtheta)$ denoting the predicted label for input $\vx_i$. 

When imposing penalty on the gradient norm of the loss function, a term with respect to it could be added on the loss function $L_{\gS}(\vtheta)$ simply,
\begin{equation}
    \label{eqn : loss}
    L(\vtheta) = L_{\gS}(\vtheta) + \lambda \cdot || \nabla_{\vtheta} L_{\gS}(\vtheta) ||_p
\end{equation}
where $||\cdot||_p$ denotes the $\normlp$-norm and $\lambda$ is the penalty coefficient and $\lambda \in \sR_{+}$ (in the experiment section, we also investigate the results where $\lambda \in \sR_{-}$). And for clarity, we would use $\normltwo$-norm ($p=2$) in the following demonstration since it is the most commonly used metric in deep learning.

\subsection{Gradient Norm and Lipschitz Continuity}
\label{sec : lip con}

Generally, penalizing the gradient norm of loss function would motivate the loss function to have small Lipschitz constant in local. If the loss function has a smaller Lipschitz constant, it would indicate that the loss function landscape is flatter, which in consequence could lead to better model generalization. 

Regarding the term "flat minima", it is a rather intuitive concept. Based on the description in \cite{DBLP:journals/neco/HochreiterS97a}, a flat minimum denotes "a large connected region in weight space where the error remains approximately constant". However, the mathematical descriptions may differ \cite{DBLP:journals/neco/HochreiterS97a,DBLP:conf/nips/NeyshaburBMS17,DBLP:conf/icml/DinhPBB17,DBLP:conf/iclr/KeskarMNST17,DBLP:conf/iclr/ChaudhariCSLBBC17}, although they may convey similar core ideas. Here, we would only follow the basic concept in our demonstration.

We would start from the Lipschitz continuous. Given $\Omega \subset \sR^n$, for function $h : \Omega \rightarrow \sR^m$, it is called Lipschitz continuous if there exists a constant $K$ that satisfies,
\begin{equation}
    \label{eqn : Lipschitz}
    ||h(\vtheta_1) - h(\vtheta_2)||_2 \leq K \cdot \Vert \vtheta_1 - \vtheta_2||_2
\end{equation}
for $\forall \vtheta_1, \vtheta_2 \in \Omega$. And the Lipschitz constant generally refers to the smallest $K$ of the function. Further, for $\forall \vtheta \in \Omega$, $h$ is locally Lipschitz continuous if $\vtheta$ has a neighborhood $\gA$ that $h|_{\gA}$ is Lipschitz continuous.

Intuitively, the Lipschitz constant describes the upper bound on the output change in the input space $\Omega$. In particular, for $h|_{\gA}$, it would indicate the supremum of output change in the neighborhood $\gA$.  In other words, for small Lipschitz constants, given any two points in $\gA$, the gap between their outputs is limited to a small range. In fact, this is essentially a kind of description of "flat minima".

So for a minimum $\vtheta_i$ and the loss function $L(\vtheta)$, according to the mean value theorem, the differentiability could lead to that for $\forall \vtheta_i^{'} \in \gA$,
\begin{equation}
    \Vert L(\vtheta_i^{'}) - L(\vtheta_i) \Vert_2 = \Vert \nabla L(\vzeta)(\vtheta_i^{'} - \vtheta_i) \Vert_2
\end{equation}
where $\vzeta = c \vtheta_i + (1 - c) \vtheta_i^{'}$, $c \in [0, 1]$. And the Cauchy-Schwarz inequality gives,
\begin{equation}
    \Vert L(\vtheta_i^{'}) - L(\vtheta_i) \Vert_2 \leq \Vert \nabla L(\vzeta) \Vert_2 \Vert (\vtheta_i^{'} - \vtheta_i) \Vert_2
\end{equation}
When $\vtheta_i^{'} \rightarrow \vtheta$, the corresponding Lipschitz constant approximates to $\Vert \nabla L(\vtheta_i) \Vert_2$. Therefore, we would expect to reduce $\Vert \nabla L(\vtheta_i) \Vert_2$ to give small Lipschitz constants such that models could converge to flat minima.

Additionally, it should be especially discriminated that some works \cite{DBLP:journals/corr/YoshidaM17,DBLP:conf/nips/VirmauxS18} try to regularize the Lipschitz constant of DNNs in the input space such that models would be more stable to the perturbation in the input space. This is not the same as penalizing the gradient norm of loss function, which would function in the weight space. 

\subsection{Gradient Calculation of Loss with Gradient Norm Penalty}

During practical optimization, we need to calculate the gradient of current loss (\Eqref{eqn : loss}),
\begin{equation}
    \label{eqn : gp loss}
    \nabla_{\vtheta} L(\vtheta) = \nabla_{\vtheta} L_{\gS}(\vtheta) + \nabla_{\vtheta} (\lambda \cdot || \nabla_{\vtheta} L_{\gS}(\vtheta) ||_p)
\end{equation}
Based on the chain rule, \Eqref{eqn : gp loss} could be simplified as,
\begin{equation}
    \label{eqn : gradient of loss}
    \nabla_{\vtheta} L(\vtheta) = \nabla_{\vtheta} L_{\gS}(\vtheta) + \lambda \cdot \nabla_{\vtheta}^2 L_{\gS}(\vtheta) \frac{\nabla_{\vtheta} L_{\gS}(\vtheta)}{||\nabla_{\vtheta} L_{\gS}(\vtheta)||} 
\end{equation}
In Appendix, we have provided detailed procedures for this simplification from \Eqref{eqn : gp loss} to \Eqref{eqn : gradient of loss}. 

Apparently, \Eqref{eqn : gradient of loss} involves the calculation of Hessian matrix. For DNNs, it is infeasible to straightforwardly solve such a Hessian matrix since the dimension in weight space is too huge. Appropriate approximation method should be implemented in this calculation.

In \Eqref{eqn : gradient of loss}, the Hessian matrix is essentially a linear operator $\mH(\cdot)$ that functions on the corresponding gradient vector. Here, local Taylor expansion would be employed to approximate the operation results between the Hessian matrix and the gradient vector. From the Taylor expansion, we have
\begin{equation}
    \label{eqn : Hessian Tylor}
    \nabla_{\vtheta}L_\gS(\vtheta + \Delta \vtheta) = \nabla_{\vtheta}L_\gS(\vtheta) + \mH\Delta \vtheta + \gO(||\Delta \vtheta||^2)
\end{equation}
When choosing $\Delta \vtheta = r\vv$ where $r$ is a small value and $\vv$ is a vector, \Eqref{eqn : Hessian Tylor} would be,
\begin{equation}
    \label{eqn : Hessian linear Op}
    \mH\vv = \frac{\nabla_{\vtheta}L_\gS(\vtheta +r\vv) - \nabla_{\vtheta}L_\gS(\vtheta)}{r} + \gO(r)
\end{equation}
Further, assigning $\vv = \frac{\nabla_{\vtheta}L_{\gS}(\vtheta)}{||\nabla_{\vtheta}L_{\gS}(\vtheta)||} $,
\begin{equation}
    \label{eqn : Hessian appro}
    \mH\frac{\nabla_{\vtheta}L_{\gS}(\vtheta)}{||\nabla_{\vtheta}L_{\gS}(\vtheta)||} \approx \frac{\nabla_{\vtheta}L(\vtheta +r\frac{\nabla_{\vtheta}L_{\gS}(\vtheta)}{||\nabla_{\vtheta}L_{\gS}(\vtheta)||}) - \nabla_{\vtheta}L(\vtheta)}{r}
\end{equation}
Now, based on \Eqref{eqn : Hessian appro}, \Eqref{eqn : gradient of loss} would be,
\begin{equation}
    \label{eqn : final loss}
    \begin{split}
    \nabla_{\vtheta} L(\vtheta) & = \nabla_{\vtheta} L_{\gS}(\vtheta) \\
    & + \frac{\lambda}{r} \cdot (\nabla_{\vtheta}L_\gS(\vtheta +r\frac{\nabla_{\vtheta}L_{\gS}(\vtheta)}{||\nabla_{\vtheta}L_{\gS}(\vtheta)||}) - \nabla_{\vtheta}L_\gS(\vtheta))  \\
    & = (1 - \alpha) \nabla_{\vtheta} L_{\gS}(\vtheta) \\
    & + \alpha \nabla_{\vtheta}L_\gS(\vtheta +r\frac{\nabla_{\vtheta}L_{\gS}(\vtheta)}{||\nabla_{\vtheta}L_{\gS}(\vtheta)||}) \\
    \end{split}
\end{equation}
where $\alpha =  \frac{\lambda}{r}$, and we would call $\alpha$ the balance coefficient. 

Accordingly, we need to set two basic parameters $\lambda$ and $r$ to perform gradient norm penalty. For $\lambda$, it denotes the penalty coefficient, which controls the degree of the regularization on the gradient norm. However, the connections between weight norm, gradient norm and model generalization is subtle during training. Currently, how much the gradient norm should be penalized in practical training still requires some further tuning effort. As for $r$, it is used for appropriating the Hessian multiplication operation (\Eqref{eqn : Hessian linear Op}). Notably, $r$ should be set carefully here since it would directly affect the approximation precision \cite{DBLP:journals/neco/Pearlmutter94}. On the one hand, $r$ is expected to be small enough such that the term $\gO(r)$ in approximation could be safely ignored. But on the other hand, as $r$ becomes smaller, the perturbed weight $\vtheta +r\vv$ will gradually weaken the effect of $\vv$ and approach the reference weight $\vtheta$, which makes $\nabla_{\vtheta}L_\gS(\vtheta +r\vv)$ and $\nabla_{\vtheta}L_\gS(\vtheta)$ too close when appropriating $\mH \vv$. Therefore, we should also avoid setting $r$ too small to provide enough precision of $\vv$.

% Based on our following experimental investigation, it is best to set $0 \leq \alpha \leq 1$.

In practice, we would further take an approximation for computing the second term in \Eqref{eqn : final loss} to avoid the Hessian computation caused by the chain rule,
\begin{equation}
    \nabla_{\vtheta}L_\gS(\vtheta +r\frac{\nabla_{\vtheta}L_{\gS}(\vtheta)}{||\nabla_{\vtheta}L_{\gS}(\vtheta)||}) \approx \nabla_{\vtheta} L_{\gS}(\vtheta)|_{\vtheta = \vtheta +r\frac{\nabla_{\vtheta}L_{\gS}(\vtheta)}{||\nabla_{\vtheta}L_{\gS}(\vtheta)||}}
\end{equation}

\begin{algorithm}[tb]
    \caption{Optimization Scheme of Penalizing Gradient Norm}
    \label{alg : gradient penalty}
    \textbf{Input}: Training set $\gS = \{(\vx_i, \vy_i)\}_{i = 0}^{N}$; loss function $L(\cdot)$; batch size $B$; learning rate $\eta$; total step $T$; balance coefficient $\alpha$; approximation scalar $r$. \\
    \textbf{Parameter}: Model parameters $\vtheta$ \\ 
    \textbf{Output}: Optimized weight $\hat{\vtheta}$
    \begin{algorithmic}[1]
    \STATE {Parameter initialization $\vtheta_{0}$.}
    \FOR{step $t=1$ {\bfseries to} $T$}
    \STATE{Get batch data pairs $\gB = \{(\vx_i, \vy_i)\}_{i = 0}^{B}$ sampled from training set $\gS$.}
    \STATE{Calculate the gradient $\vg_1 = \nabla_{\vtheta}L_\gS(\vtheta)$ based on the batch samples.}
    \STATE{Add $r\frac{\nabla_{\vtheta}L_{\gS}(\vtheta)}{||\nabla_{\vtheta}L_{\gS}(\vtheta)||}$ on the current parameter $\vtheta_t$, which makes $\vtheta_t{'} = \vtheta_t + r\frac{\nabla_{\vtheta}L_{\gS}(\vtheta)}{||\nabla_{\vtheta}L_{\gS}(\vtheta)||}$.}
    \STATE{Calculate the gradient $\vg_2 = \nabla_{\vtheta}L_{\gS}(\vtheta)$ at $\vtheta = \vtheta_t{'}$.}
    \STATE{Calculate the final gradient $\vg =  (1-\alpha)\vg_1 + \alpha \vg_2$.}
    \STATE{(SGD optimizer) Update parameter with final gradient, $\vtheta_{k + 1} = \vtheta_{k} - \eta \cdot \vg$.}
    \ENDFOR
    \STATE{\textbf{return} Final optimization $\hat{\vtheta}$}.
    \end{algorithmic}
\end{algorithm}

\begin{table*}[tb]
    \centering
    \caption{Testing error rate of CNN models on Cifar10 and Cifar100 when implementing the three training schemes.}
    \vskip 0.15in
    \begin{tabular}{lcccc}
    \toprule
     & \multicolumn{2}{|c|}{Cifar10} & \multicolumn{2}{|c|}{Cifar100} \\
    \midrule
    VGG16 & \multicolumn{1}{|c}{Basic} & \multicolumn{1}{c|}{Cutout} & \multicolumn{1}{|c}{Basic} & \multicolumn{1}{c|}{Cutout} \\
    \midrule
    Standard & $\ 7.07\ $ & $\ 5.31\ $ & $\ 28.78\ $  & $\ \textcolor{red}{\mathbf{26.98}}\ $  \\
    SAM & $\ 6.91\ $ & $\ \textcolor{red}{\mathbf{6.17}} \ $ & $\ 28.62\ $  & $\ 27.13\ $  \\
    \textbf{Ours} & $\ \ \mathbf{6.72}\ \ $ & $\ \  \mathbf{5.19} \ \ $ & $\ \ \mathbf{28.48}\ \ $  & $\ \ 27.07\ \ $ \\
    \midrule
    VGG16-BN & \multicolumn{1}{|c}{Basic} & \multicolumn{1}{c|}{Cutout} & \multicolumn{1}{|c}{Basic} & \multicolumn{1}{c|}{Cutout} \\
    \midrule
    Standard & $\ 5.74_{ \pm 0.09}\ $ & $\ 4.39_{ \pm 0.07}\ $ & $\ 25.22_{ \pm 0.31}\ $  & $\ 24.69_{ \pm 0.25}\ $  \\
    SAM & $\ 5.24_{ \pm 0.08}\ $ & $\ 4.16_{ \pm 0.11}\ $ & $\ 24.23_{ \pm 0.29}\ $  & $\ 23.35_{ \pm 0.33}\ $  \\
    \textbf{Ours} & $\ \ \mathbf{4.88_{ \pm 0.12}}\ \ $ & $\ \  \mathbf{4.02_{ \pm 0.08}}\ \ $ & $\ \ \mathbf{24.04_{ \pm 0.18}}\ \ $  & $\ \ \mathbf{23.07_{ \pm 0.26}}\ \ $ \\
    \midrule
    WideResNet-28-10\ \ \ \  & \multicolumn{1}{|c}{Basic} & \multicolumn{1}{c|}{Cutout} & \multicolumn{1}{|c}{Basic} & \multicolumn{1}{c|}{Cutout} \\
    \midrule
    Standard & $\ 3.53_{ \pm 0.10}\ $ & $\ 2.81_{ \pm 0.07}\ $ & $\ 18.99_{ \pm 0.12}\ $  & $\ 16.92_{ \pm 0.10}\ $  \\
    SAM & $\ 2.78_{ \pm 0.07}\ $ & $\ 2.43_{ \pm 0.13}\ $ & $\ 16.53_{ \pm 0.13}\ $  & $\ 14.87_{ \pm 0.16}\ $  \\
    \textbf{Ours} & $\ \ \mathbf{2.52_{ \pm 0.09}}\ \ $ & $\ \  \mathbf{2.16_{ \pm 0.11}}\ \ $ & $\ \ \mathbf{16.02_{ \pm 0.19}}\ \ $  & $\ \ \mathbf{14.28_{ \pm 0.16}}\ \ $ \\
    \midrule
    WideResNet-SS 2$\times$96  & \multicolumn{1}{|c}{Basic} & \multicolumn{1}{c|}{Cutout} & \multicolumn{1}{|c}{Basic} & \multicolumn{1}{c|}{Cutout} \\
    \midrule
    Standard & $\ 2.82_{ \pm 0.05}\ $ & $\ 2.39_{ \pm 0.06}\ $ & $\ 17.19_{ \pm 0.19}\ $  & $\ 15.85_{ \pm 0.14}\ $  \\
    SAM & $\ 2.37_{ \pm 0.09}\ $ & $\ 2.11_{ \pm 0.13}\ $ & $\ 15.22_{ \pm 0.19}\ $  & $\ 14.32_{ \pm 0.15}\ $  \\
    \textbf{Ours} & $\ \ \mathbf{2.28_{ \pm 0.13}}\ \ $ & $\ \  \mathbf{2.01_{ \pm 0.10}}\ \ $ & $\ \ \mathbf{14.93_{ \pm 0.10}}\ \ $  & $\ \ \mathbf{14.03_{ \pm 0.17}}\ \ $ \\
    \midrule
    PyramidNet-SD & \multicolumn{2}{|c|}{Auto Aug + Cutmix} & \multicolumn{2}{|c|}{Auto Aug + Cutmix} \\
    \midrule
    Standard & \multicolumn{2}{c}{$1.66_{ \pm 0.11}$} & \multicolumn{2}{c}{$10.83_{ \pm 0.14}$}  \\
    SAM & \multicolumn{2}{c}{$1.41_{ \pm 0.08}$} & \multicolumn{2}{c}{$10.33_{ \pm 0.13}$}  \\
    \textbf{Ours} & \multicolumn{2}{c}{$\mathbf{1.30_{ \pm 0.07}}$}  & \multicolumn{2}{c}{$\mathbf{10.12_{ \pm 0.17}}$}  \\
    % \midrule
    % ViT-B16 & \multicolumn{1}{|c}{Basic} & \multicolumn{1}{c|}{Cutout} & \multicolumn{1}{|c}{Basic} & \multicolumn{1}{c|}{Cutout} \\
    % \midrule
    % Standard & $\ 80.04_{ \pm 0.11}\ $ & $\ 82.73_{ \pm 0.21}\ $ & $\ 54.42_{ \pm 0.25}\ $  & $\ 57.79_{ \pm 0.20}\ $  \\
    % SAM & $\ 80.01_{ \pm 0.18}\ $ & $\ 82.73_{ \pm 0.17}\ $ & $\ 54.61_{ \pm 0.31}\ $  & $\ 57.81_{ \pm 0.17}\ $  \\
    % \textbf{Ours} & $\ \ \mathbf{81.11_{ \pm 0.12}}\ \ $ & $\ \  \mathbf{84.03_{ \pm 0.16}}\ \ $ & $\ \ \mathbf{55.76_{ \pm 0.20}}\ \ $  & $\ \ \mathbf{59.52_{ \pm 0.22}}\ \ $ \\
    \bottomrule
    \end{tabular}
    \label{tbl : cifar}
\end{table*}

In summary, \Algref{alg : gradient penalty} gives the full procedures of our optimization scheme. It should be mentioned that \Algref{alg : gradient penalty} only shows our scheme when using SGD optimizer. For other optimizers or update strategies (such as Adam), one could add specific operations before step 8. 

% \subsection{Discussions}

% \paragraph{Deployment of parameters} According to \Eqref{eqn : final loss}, we need to set two basic parameters $\lambda$ and $r$ to perform the gradient norm penalty in practice. For $\lambda$, it denotes the penalty coefficient, which controls the degree of the regularization on the gradient norm. However, the connections between weight norm, gradient norm and the model performance is subtle during training. Currently, how much the gradient norm should be penalized in practical training still requires some further tuning effort. As for $r$, it is used for appropriating the Hessian multiplication operation (\Eqref{eqn : Hessian linear Op}). Notably, $r$ should be set carefully here since it would directly affect the approximation precision \cite{DBLP:journals/neco/Pearlmutter94}. On the one hand, $r$ is expected to be small enough such that the term $\gO(r)$ in approximation could be safely ignored. But on the other hand, as $r$ becomes smaller, 

% $r$ should not be too small in order to provide valid gap between $\vtheta + r\vv$ and $\vtheta$. Otherwise, using Eq 8 to compute Hv would 

% $r$ should not be set too small to provide enough precision of approximation $\mH \vv$. 

Particularly, if $\alpha = 1$, it is actually the SAM optimization \cite{DBLP:conf/iclr/ForetKMN21}. This indicates that SAM is a special implementation of penalizing gradient norm, where the penalty coefficient $\lambda$ is always set equal to $r$. However, given the distinct roles of the two parameters, we can not anticipate that models would achieve best performance every time at $\lambda = r$. Such a binding deployment, while reducing one parameter, also limits our tuning. We would further show that SAM could not be the best implementation in the following experiment section.

% It should be noted that $\lambda$ denotes the penalty coefficient. It controls how much we would like to penalize the gradient norm. But like L2 norm regularization, how much we should implement is still unclear in theory, requiring some tuning effort. For $r$, we should also carefully implement in practice. In general, we would expect to set $r$ to be smaller enough such that the $Or$ could be safely neglected. But on the other hand, $r$ should not be set too small to provide enough precision of approximation $Hv$. 

\section{Experiments}
We would demonstrate the effectiveness of our proposed scheme by investigating performance on image classification tasks. In all our experiments, we would compare our scheme with two other training schemes, one is the standard training scheme and the other one is the SAM training scheme. Besides, all the experiments are deployed using the JAX framework on the NVIDIA DGX Station A100. 

\subsection{Cifar10 and Cifar100}
In this section, we would use Cifar10 and Cifar100 as our experimental datasets, and would separately apply the convolutional neural network (CNN) architectures and the vision transformer (ViT) architecture for the corresponding benchmark performance tests.

\paragraph{Convolutional neural network}

For CNN architectures, five different architectures would be involved, including relatively simple architectures (VGG16 \cite{DBLP:journals/corr/SimonyanZ14a}) and complex architectures (WideResNet \cite{DBLP:conf/bmvc/ZagoruykoK16} and PyramidNet \cite{DBLP:conf/cvpr/HanKK17}).

For training datasets, we would employ two kinds of augmentations. The first one is the basic augmentation, where samples are padded with additional four pixels on each boundary, randomly flipped in horizontal and then cropped randomly to size $32 \times 32$. The second one is the cutout augmentation, where cutout regularization \cite{DBLP:journals/corr/abs-1708-04552} would be implemented moreover based on the basic augmentation. Specifically, cutout regularization is a common data augmentation technique for CNNs, which randomly masks out a square region in the input with a given mask value (generally zero). 

Our investigation would focus on the comparisons between three different training schemes, namely the standard SGD scheme, SAM scheme and our scheme. Considering the connections between these three training schemes, we would adopt a "greedy" strategy to reduce tuning cost during implementation,
\begin{enumerate}
    \item We would first train models using the standard training scheme ($\lambda = 0$ in our scheme), and record the common training hyperparameters (such as learning rate, weight decay) that could give the best model performance.
    \item Based on the best hyperparameters recorded in Step 1, we would then set the balance coefficient $\alpha = 1$ to train the same models using the SAM scheme, and perform a grid search on the parameter $r$.
    \item Finally, based on the previous information, we would further perform a grid search on the balance coefficient $\alpha$ to adjust the penalty coefficient. 
\end{enumerate}

Basically, the involved model architectures have been extensively studied for the standard training scheme. It is not necessary to perform a heavy grid search, and we could just follow the common hyperparameters used in the related literatures. Next, we would perform a grid search on the scaler $r$ over the set $\{0.01, 0.02, 0.05, 0.1, 0.2\}$. This setting is actually the same as it in \cite{DBLP:conf/iclr/ForetKMN21}. And for a fair comparison, we would train the models to reach at comparable results as reported in their paper. After determining the best value of $r$, we would moreover perform a grid search on the balance coefficient $\alpha$ in the range 0.1 to 0.9 at an interval of 0.1.

For each model, we would train with five different random seeds, and record the convergence model performance on testing sets during training. And then we would report the mean value and the standard deviation, as shown in Table \ref{tbl : cifar}. 

In table \ref{tbl : cifar}, totally five model cases are involved: the original VGG16 architecture and that with batch normalization regularization, the WideResNet-28-10 architecture and that with Shake-Shake regularization \cite{DBLP:conf/iclr/Gastaldi17}, and the PyramidNet-272 architecture with Shake-Drop regularization \cite{2019/shakedrop}. We could see that for our training scheme, the model performance could be improved to some extent compared to the other two schemes.

For the original VGG16 model, training could frequently fail with relatively large learning rate, sometimes all five trials may fail especially for the SAM scheme. However, small learning rate may not lead to the best performance. Here, we would simply report the best training case (possibly trained more than five times and not following the greed strategy strictly). 

Basically, the results are quite close when using the three training schemes. But it should be highly noted about the results when implementing cutout augmentation. On Cifar10, the performance using SAM scheme may be much worse than that using standard scheme. This is because that for the optimal common training hyperparameters in the standard scheme, all of the values in our grid search on approximation scalar $r$ fail to train from the start using SAM scheme. We have to lower the learning rate to stabilize the training. In contrast, by setting appropriate balance coefficient $\alpha$, our scheme could utilize the optimal common training hyperparameters, which could improve the performance slightly. However, on Cifar100, the standard training would yield the best performance. 

When applying the batch normalization regularization on the VGG model, such training failures would be largely alleviated, although they may still happen in few trials. We would see that with batch normalization, VGG16 could receive significantly performance gains. The best performances are achieved via our scheme, which could be as low as 4\% testing error rate on Cifar10.

As for the WideResNet architecture, we could find that our scheme could significantly improve the performance by 1\% on Cifar10 and near 3\% on Cifar100 compared to the standard training scheme. Our improvements could be 38\% on average and up to remarkable 70\% (on Cifar10 with cutout, ours is 0.65 while SAM's is 0.38, which is $0.65 = (1 + 71\%) \times 0.38$) more than the SAM's improvements over the standard training scheme. This confirms the effectiveness of our scheme, and further demonstrates that SAM is not the best case in our scheme.

Regarding the WideResNet with Shake-Shake regularization (WideResNet-SS in the table), our improvements could not be as significant as that on the WideResNet architecture, but still give about 20\% improvement compared to the SAM's improvement over the standard training scheme.

\begin{figure*}[!htb]
    \centering
    \center{\includegraphics[width = 1.6\columnwidth]{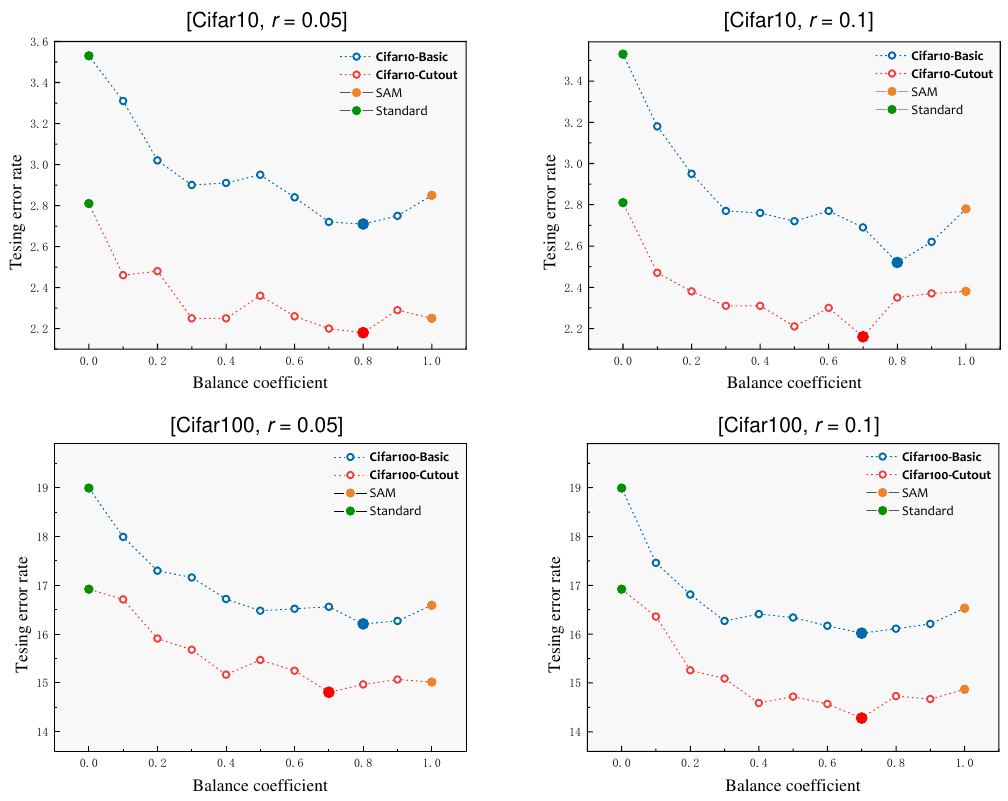}}
    \caption{Testing error rate when trained with different hyperparameter $r$ and $\alpha$. The upper row denotes the error rate on Cifar10 while the lower row denotes the error rate on Cifar100, and the left column denotes $r = 0.05$ while the right column denotes $r = 0.1$. The points colored green are results using the standard scheme, and the points colored orange are results using the SAM scheme.}
    \label{fig: study of parameters}
\end{figure*}

Finally, we would investigate the PyramidNet architecture with the Shake-Drop regularization (PyramidNet-SD in the table). We would adopt the auto-augmentation policy \cite{DBLP:journals/corr/abs-1805-09501} and the cutmix regularization \cite{DBLP:conf/iccv/YunHCOYC19} for data augmentation. Here, only three random seeds are used for training. We could see that our scheme again improves the performance on both Cifar10 and Cifar100.

\paragraph{Vision transformer}

We would like to investigate the effectiveness of our scheme on the recent vision transformer architectures \cite{DBLP:conf/iclr/DosovitskiyB0WZ21}. Our investigation would focus on the ViT-TI16 and ViT-S16 architectures introduced in their paper.

\begin{table}[t]
    \caption{Testing error rate of ViT models on Cifar10 dataset when implementing the three training schemes.}
    \vskip 0.15in
    \centering
    \begin{tabular}{lcc}
        \toprule
        & \multicolumn{2}{|c|}{Cifar10} \\
        \midrule
        ViT-TI16 & \multicolumn{1}{|c}{Basic} & \multicolumn{1}{c|}{Heavy} \\
        \midrule
        Standard & $\ 15.92_{ \pm 0.17}\ $ & $\ 14.68_{ \pm 0.14}\ $   \\
        SAM & $\ 15.33_{ \pm 0.18}\ $ & $\ 13.77_{ \pm 0.12}\ $ \\
        Ours & $\ \mathbf{14.75_{ \pm 0.17}}\ $ & $\ \mathbf{13.52_{ \pm 0.21}}\ $ \\
        \midrule
        ViT-S16 \ \ \ & \multicolumn{1}{|c}{Basic} & \multicolumn{1}{c|}{Heavy} \\
        \midrule
        Standard & $\ 14.55_{ \pm 0.14}\ $ & $\ 13.31_{ \pm 0.11}\ $   \\
        SAM & $\ 13.91_{ \pm 0.18}\ $ & $ \ 12.63_{ \pm 0.09}\ $ \\
        Ours & $\ \mathbf{13.66_{ \pm 0.16}}\ $ & $\ \mathbf{12.29_{ \pm 0.19}}\ $ \\
        \bottomrule
    \end{tabular}
    % \vskip 0.1in
    \label{tbl : vit cifar10}
\end{table}

\begin{table}[ht]
    \caption{Testing error rate of ViT models on Cifar100 dataset when implementing the three training schemes.}
    \vskip 0.15in
    \centering
    \begin{tabular}{lcc}
        \toprule
        & \multicolumn{2}{|c|}{Cifar100} \\
        \midrule
        ViT-TI16 & \multicolumn{1}{|c}{Basic} & \multicolumn{1}{c|}{Heavy} \\
        \midrule
        Standard & $\ 40.21_{ \pm 0.20}\ $ & $\ 38.93_{ \pm 0.28}\ $   \\
        SAM & $\ 38.89_{ \pm 0.23}\ $ & $\ 37.61_{ \pm 0.19}\ $ \\
        Ours & $\ \mathbf{38.58_{ \pm 0.27}}\ $ & $\ \mathbf{37.15_{ \pm 0.21}}\ $ \\
        \midrule
        ViT-S16 \ \ \ & \multicolumn{1}{|c}{Basic} & \multicolumn{1}{c|}{Heavy} \\
        \midrule
        Standard & $\ 38.43_{ \pm 0.19}\ $ & $\ 37.58_{ \pm 0.22}\ $   \\
        SAM & $\ 37.98_{ \pm 0.23}\ $ & $ \ 36.77_{ \pm 0.25}\ $ \\
        Ours & $\ \mathbf{37.32_{ \pm 0.28}}\ $ & $\ \mathbf{36.59_{ \pm 0.22}}\ $ \\
        \bottomrule
    \end{tabular}
    % \vskip 0.1in
    \label{tbl : vit cifar100}
\end{table}

\begin{figure*}[ht]
    \vskip 0.2in
    \centering
    \center{\includegraphics[width = 1.6\columnwidth]{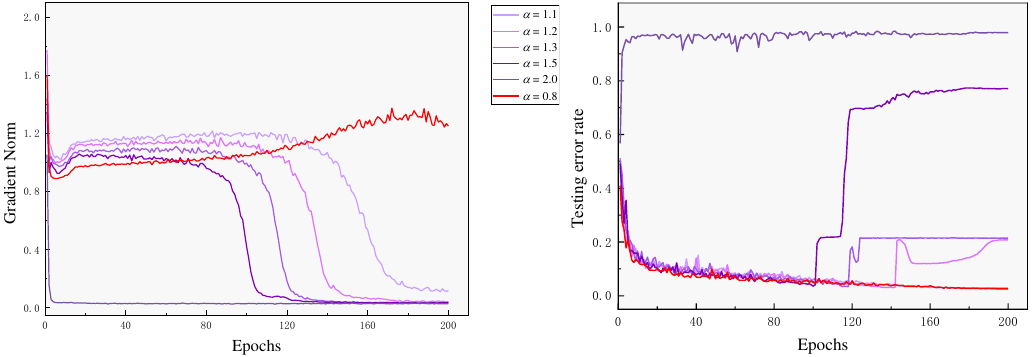}}
    \caption{Evolutions of gradient norm of loss function (left) and testing error rate on Cifar10. Purple curves represent the results where $\alpha \in \gB_b$, while the red curve represents the optimal results where $\alpha = 0.8$.}
    \label{fig: study of set B}
\end{figure*}

Here, we would adopt the same greedy search strategy as in the previous section for the three training schemes. Likewise, five random seeds are used for each model. But for data augmentation, we would not use cutout augmentation here since we find such augmentation would not boost the model performance. Intuitively, the vision transformer architecture would cut the image into small patches, and utilize the relationship between these patches to make decisions. In this way, the cutout regularization may not be helpful for models to learn the relationship between patches. Here, we would replace the cutout augmentation with a heavy augmentation, considering that vision transformer architectures are generally data hungry models. In the heavy augmentation, we would perform a series of operations, including resizing to $72 \times 72$, random flipping, random rotating, random zooming, random cropping and finally resizing to $48 \times 48$. We would adopt the $4 \times 4$ patch size in both the basic augmentation and the heavy augmentation. Models would be trained much longer using heavy augmentation than those using basic augmentation (1200 v.s 300 epochs). In addition, we would use extra operations like label smoothing and drop path as used in \cite{DBLP:conf/iclr/DosovitskiyB0WZ21} when using the heavy augmentation.

Table \ref{tbl : vit cifar10} \& \ref{tbl : vit cifar100} presents the corresponding testing error rate. We could see that even if using heavy augmentation, the performances of vision transformer architectures would be much worse than those of CNN architectures. And in the table, we could find that the performances could be improved via implementing our scheme. This further confirms the broad applicability of our scheme.

\paragraph{Parameter study}
Further, we would investigate the impact on model performance as choosing different balance coefficients $\alpha$ and approximation scalars $r$ in our optimization scheme, which we would illustrate using the WideResNet-28-10 model architecture. 

When performing the grid search on approximate value $r$ in the SAM training experiments, we observe that models would have relatively better performances when setting $r = 0.05$ and $r = 0.1$. This observation is the same as that in \cite{DBLP:conf/iclr/ForetKMN21}. Then, the grid search over balance coefficient $\alpha$ is performed moreover based on $r = 0.05$ and $0.1$. \Figref{fig: study of parameters} shows the results. In \Figref{fig: study of parameters}, rows denote results on Cifar10 and Cifar100, respectively, and blue lines in the plots denote adopting basic data augmentation while red lines denote adopting cutout data augmentation. As we would see in the figure, from the standard training scheme ($\alpha = 0$) to the SAM training scheme ($\alpha = 1$), each curve may experience a decrease and then an increase in the testing error rate. Based on the figure, we could find that these models would achieve best performance when the balance coefficient $\alpha$ is set around 0.7 or 0.8.

% From previous demonstrations, $r$ should not be set too large or too small to keep the approximation precise. 

In addition to the basic set in the previous deployment of $\alpha$, we would like to further investigate cases where $\alpha \notin [0, 1]$. Extra deployments would be implemented over two other sets, $\gB_a = \{-0.1, -0.2, -0.5\}$ and $\gB_b = \{1.1, 1.2, 1.3, 1.5, 2.0\}$. 

% Here, the same greedy strategy is still used for the three training schemes. Also, the same grid search would performed on approximate value $r$ as in the previous section. For the balance coefficient $\alpha$, in addition to the basic set in the previous section, we would perform extra search over two other sets, $\gB_a = \{-0.1, -0.2, -0.5\}$ and $\gB_b = \{1.1, 1.2, 1.3, 1.5, 2.0\}$. 

% When implementing the grid search on approximate value $r$, we observe that models would have relatively better performances when setting $r = 0.05$ and $r = 0.1$. This observation is the same as that in \cite{DBLP:conf/iclr/ForetKMN21}. Then, the grid search over the basic set of balance coefficient $\alpha$ is performed moreover. \Figref{fig: study of parameters} shows the final results. In \Figref{fig: study of parameters}, rows denote results on Cifar10 and Cifar100, respectively, and blue lines in the plots denote adopting basic data augmentation while red lines denote adopting cutout data augmentation. As we would see in the figure, from the standard training scheme ($\alpha = 0$) to the SAM training scheme ($\alpha = 1$), each curve may experience a decrease and then an increase in the testing error rate. Based on the figure, we could find that these models would achieve best performance when the balance coefficient $\alpha$ is set around 0.8.

For $\gB_a$, since its values are all negative, this causes that the penalty coefficient of gradient norm in \Eqref{alg : gradient penalty} becomes negative, which makes it a reward as increasing the gradient norm during optimization. Generally, if $\lambda \leq 0$, the optimization would not be fully ensured, since we are adding a negative term on the loss. And in all of our trials, no matter trained on Cifar10 or Cifar100 datasets, the models completely fail to converge even if gradient clip regularization is adopted. The gradient would be unstable, and may even explode immediately after the training start. But this instead shows the effectiveness of our penalty scheme.

As for $\gB_b$, since the values in it are all greater than 1.0, the penalty on the gradient norm becomes larger, and the operation relationship in \Eqref{eqn : final loss} shifts from addition to subtraction. This may somehow be harmful to training, as shown in \figref{fig: study of set B}. In the figure, the left plot denotes the gradient norm of loss function with respect to the training epochs, while the right plot denotes the testing error rate on Cifar10 datasets.

We could find that these large coefficients indeed impose much heavier penalties on the gradient norm of loss function. The gradient norm would drop faster and faster as $\alpha$ increases from 1.1 to 2.0. For $\alpha = 2.0$, the gradient norm would drop to near zero immediately after the training start. As for other values in $\gB_b$, although the gradient norm would keep stable for a while, it would suddenly drop rapidly within only several epochs. As soon as the gradient norm begins to drop rapidly, the testing error rate would increase immediately. When the gradient norms reach near zero, the testing error rates become stable. Interestingly, the final convergence error rates may be different even though the corresponding gradient norms are all near zero. 

In summary, one should impose the penalty on the gradient norm with appropriate parameters in practice, where $\alpha = 0.8$ and $r \in \{0.05, 0.1\}$ are highly recommended for achieving the best performance. Based on our observation, this deployment could also give the best performance for most of our training, not just the WideResNet-28-10 architecture.

\subsection{ImageNet}
Next, we would check the effectiveness of our scheme on the large-scale dataset, namely ImageNet. For model architectures, we would adopt the VGG16-BN, ResNet50 and ResNet101 in our investigation. Likewise, we would still adopt the three training schemes for comparisons. However, unlike using the greedy strategy for hyperparameter searching in the previous section, we would directly set $r = 0.05$ according to \cite{DBLP:conf/iclr/ForetKMN21} and perform only a slight grid search on $\alpha$ over $\{0.7, 0.8\}$ based on our parameter study. For data augmentation, we just follow the prior works \cite{DBLP:conf/cvpr/HeZRS16,DBLP:journals/corr/SimonyanZ14a}. Here, we would train each model with three different random seeds. Besides, all models are trained within 100 epochs with a cosine learning rate schedule.

\begin{table}[httb]
    \caption{Testing error rate of models on ImageNet dataset when implementing the three training schemes.}
    \vskip 0.15in
    \centering
    \begin{tabular}{lcc}
        \toprule
        & \multicolumn{2}{|c|}{ImageNet} \\
        \midrule
        VGG16-BN & \multicolumn{1}{|c}{Top-1 Accuracy} & \multicolumn{1}{c|}{Top-5 Accuracy} \\
        \midrule
        Standard & $\ 26.89_{ \pm 0.12}\ $ & $\ 8.88_{ \pm 0.06}\ $   \\
        SAM & $\ 26.41_{ \pm 0.13}\ $ & $\ 8.60_{ \pm 0.05}\ $ \\
        Ours & $\ \mathbf{26.12_{ \pm 0.16}}\ $ & $\ \mathbf{8.44_{ \pm 0.06}}\ $ \\
        \midrule
        ResNet50 \ \ \ & \multicolumn{1}{|c}{Top-1 Accuracy} & \multicolumn{1}{c|}{Top-5 Accuracy} \\
        \midrule
        Standard & $\ 23.64_{ \pm 0.17}\ $ & $\ 7.01_{ \pm 0.09}\ $   \\
        SAM & $\ 23.16_{ \pm 0.11}\ $ & $ \ 6.72_{ \pm 0.06}\ $ \\
        Ours & $\ \mathbf{22.87_{ \pm 0.15}}\ $ & $\ \mathbf{6.59_{ \pm 0.11}}\ $ \\
        \midrule
        ResNet101 \ \ \ & \multicolumn{1}{|c}{Top-1 Accuracy} & \multicolumn{1}{c|}{Top-5 Accuracy} \\
        \midrule
        Standard & $\ 21.97_{ \pm 0.09}\ $ & $\ 6.11_{ \pm 0.07}\ $   \\
        SAM & $\ 21.02_{ \pm 0.10}\ $ & $\ 5.31_{ \pm 0.09}\ $ \\
        Ours & $\ \mathbf{20.53_{ \pm 0.13}}\ $ & $\ \mathbf{5.18_{ \pm 0.08}}\ $ \\
        \bottomrule
    \end{tabular}
    \vskip 0.1in
    \label{tbl : imagenet}
\end{table}

Table \ref{tbl : imagenet} reports top-1 and top-5 testing error rates for different models. As we could see in the table, the model generalization could be improved when using our scheme compared to the other two schemes. Again, this confirms the effectiveness of our scheme for practical training.

\section{Conclusion}

In this paper, we introduce an effective scheme for penalizing the gradient norm of loss function during training optimization. In our scheme, no Hessian computation would be involved, making it efficient to be implemented in practical optimization. We confirm the effectiveness of our training scheme via image classification experiments which involve extensive model architecture on commonly used datasets. By comparing with two baselines (the standard training scheme and SAM scheme) on Cifar and ImageNet dataset, we show the superiority of our scheme, where several new state-of-art performances are achieved. Remarkably, the improvement of using our scheme may be at most 70\% greater than that of using the SAM scheme. Also, we perform a parameter study to guide the setting of optimal hyperparameters in practice. It is shown that one should carefully set the parameters, in case of losing precision of approximation during penalty.

\section*{Acknowledgements}

We would like to thank all the reviewers and the meta-reviewer for their helpful comments and kindly advices. We would like to thank Yuhan Li and Chuncheng Zhao from Intelligence Sensing Lab at Tsinghua University for the discussions.

\bibliography{main}
\bibliographystyle{icml2022}

%%%%%%%%%%%%%%%%%%%%%%%%%%%%%%%%%%%%%%%%%%%%%%%%%%%%%%%%%%%%%%%%%%%%%%%%%%%%%%%
%%%%%%%%%%%%%%%%%%%%%%%%%%%%%%%%%%%%%%%%%%%%%%%%%%%%%%%%%%%%%%%%%%%%%%%%%%%%%%%
% APPENDIX
%%%%%%%%%%%%%%%%%%%%%%%%%%%%%%%%%%%%%%%%%%%%%%%%%%%%%%%%%%%%%%%%%%%%%%%%%%%%%%%
%%%%%%%%%%%%%%%%%%%%%%%%%%%%%%%%%%%%%%%%%%%%%%%%%%%%%%%%%%%%%%%%%%%%%%%%%%%%%%%
\newpage
\appendix
\onecolumn

% You can have as much text here as you want. The main body must be at most $8$ pages long.
% For the final version, one more page can be added.
% If you want, you can use an appendix like this one, even using the one-column format.
% %%%%%%%%%%%%%%%%%%%%%%%%%%%%%%%%%%%%%%%%%%%%%%%%%%%%%%%%%%%%%%%%%%%%%%%%%%%%%%%
% %%%%%%%%%%%%%%%%%%%%%%%%%%%%%%%%%%%%%%%%%%%%%%%%%%%%%%%%%%%%%%%%%%%%%%%%%%%%%%%

\section{Simplification Process of \Eqref{eqn : gradient of loss}}

The \Eqref{eqn : gradient of loss} is,
\begin{equation}
    \nabla_{\vtheta} L(\vtheta) = \nabla_{\vtheta} L_{\gS}(\vtheta) + \nabla_{\vtheta} (\lambda \cdot || \nabla_{\vtheta} L_{\gS}(\vtheta) ||_p)
\end{equation}
\noindent where would like to simply the second term $\nabla_{\vtheta} (\lambda \cdot || \nabla_{\vtheta} L_{\gS}(\vtheta) ||_2)$.

For $\vtheta = [\theta_1, \theta_2, \cdots, \theta_n]^{\intercal}$, the 2-norm function is,
\begin{equation}
    g(\vtheta) := ||\vtheta||_2 = \sqrt{\theta_1^2 + \theta_2^2 + \cdots + \theta_n^2}
\end{equation}
The partial derivative of $g(\vtheta)$ with respect to $\theta_i$ denotes,
\begin{equation}
    \frac{\partial g(\vtheta)}{\partial \theta_i} = \frac{\theta_i}{\sqrt{\theta_1^2 + \theta_2^2 + \cdots + \theta_n^2}} = \frac{\theta_i}{||\vtheta||_2} = \frac{\theta_i}{g(\vtheta)}
\end{equation}
Therefore,
\begin{equation}
    \nabla_{\vtheta}g(\vtheta) = [\frac{\theta_1}{g(\vtheta)},  \frac{\theta_2}{g(\vtheta)},  \cdots, \frac{\theta_n}{g(\vtheta)}]^{\intercal}
\end{equation}
The gradient of $\vtheta$ denotes $h(\vtheta) := \nabla_{\vtheta}L(\vtheta)$. And the term $\nabla_{\vtheta}(||\nabla_{\vtheta}L(\vtheta)||_2)$ could be simplified as,
\begin{equation}
    \label{eqn : simplification process}
    \begin{split}
     \nabla_{\vtheta}(||\nabla_{\vtheta}L(\vtheta)||_2) & = \nabla_{\vtheta}(g \circ h)(\vtheta) \\
     & = (\nabla_{\vtheta}g(\vzeta)|_{\vzeta = h(\vtheta)}) \cdot (\nabla_{\vtheta}h(\vtheta)) \\
     & = (\frac{\vzeta}{g(\vzeta)}|_{\vzeta = h(\vtheta)}) \cdot (\nabla_{\vtheta}h(\vtheta)) \\
     & = (\frac{\nabla_{\vtheta}L(\vtheta)}{|| \nabla_{\vtheta}L(\vtheta) ||_2})  \cdot (\nabla^2 L(\vtheta)) \\
     & = \frac{1}{||\nabla_{\vtheta}L(\vtheta) ||_2} \cdot \nabla^2 L(\vtheta) \cdot \nabla_{\vtheta}L(\vtheta)
    \end{split}
\end{equation}

Back to \Eqref{eqn : gradient of loss}, the equation could be simplified based on \Eqref{eqn : simplification process},
\begin{equation}
    \begin{split}
        \nabla_{\vtheta} L(\vtheta) & = \nabla_{\vtheta} L_{\gS}(\vtheta) + \nabla_{\vtheta} (\lambda \cdot || \nabla_{\vtheta} L_{\gS}(\vtheta) ||_p)  \\
        & = \nabla_{\vtheta} L_{\gS}(\vtheta) + \lambda \cdot \nabla^2 L_{\gS}(\vtheta) \cdot \frac{\nabla_{\vtheta}L_{\gS}(\vtheta)}{|| \nabla_{\vtheta}L_{\gS}(\vtheta) ||_2}
    \end{split}
\end{equation}

\end{document}

%% file: math_commands.tex
\usepackage{amsmath,amsfonts,bm}

\def\figref#1{figure~\ref{#1}}
\def\Figref#1{Figure~\ref{#1}}

\def\Secref#1{Section~\ref{#1}}
\def\eqref#1{equation~\ref{#1}}
\def\Eqref#1{Equation~\ref{#1}}
\def\Algref#1{Algorithm~\ref{#1}}

\def\1{\bm{1}}

\def\vtheta{{\bm{\theta}}}
\def\vzeta{{\bm{\zeta}}}

\def\vg{{\bm{g}}}

\def\vv{{\bm{v}}}

\def\vx{{\bm{x}}}
\def\vy{{\bm{y}}}

\def\mH{{\bm{H}}}

\DeclareMathAlphabet{\mathsfit}{\encodingdefault}{\sfdefault}{m}{sl}
\SetMathAlphabet{\mathsfit}{bold}{\encodingdefault}{\sfdefault}{bx}{n}

\def\gA{{\mathcal{A}}}
\def\gB{{\mathcal{B}}}

\def\gO{{\mathcal{O}}}

\def\gS{{\mathcal{S}}}

\def\sR{{\mathbb{R}}}

\newcommand{\normltwo}{L^2}
\newcommand{\normlp}{L^p}